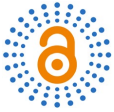

# Predicting Breast Cancer Survival: A Survival Analysis Approach Using Log Odds and Clinical Variables

Opeyemi Sheu Alamu[1], Bismar Jorge Gutierrez Choque[2], Syed Wajeeh Abbs Rizvi[3], Samah Badr Hammed[4], Isameldin Elamin Medani[5], Md Kamrul Siam[6], Waqar Ahmad Tahir[7]

[1]Departmet of Statistics, Federal College of Animal Health and Production Technology, Ibadan, Nigeria
[2]Department of Science, Universidad Mayor de San Andres, La Paz, Bolivia
[3]Department of Data Science, National University of Computer & Emerging Sciences-FAST City Campus, Karachi, Pakistan
[4]Department of Basic Medical Science, University of Khartoum, Khartoum, Sudan
[5]Consultant Obstetrics and Gynecology, University Hospital at the University of Jazan, Jazan, Saudi Arabia
[6]New York Institute of Technology, New York, USA
[7]Department of Internal Medicine/Casualty, Khalifa Gull Nawaz Teaching Hospital, Bannu, Pakistan
Email: sheuopeyemi99@gmail.com, thebismar12@hotmail.com, rizviwajih@gmail.com, samah.hamad@uofk.edu., Isameldin2015@gmail.com, ksiam01@nyit.edu, wikimarwat2641@gmail.com

## Abstract

Breast cancer remains a significant global health challenge, with prognosis and treatment decisions largely dependent on clinical characteristics. Accurate prediction of patient outcomes is crucial for personalized treatment strategies. This study employs survival analysis techniques, including Cox proportional hazards and parametric survival models, to enhance the prediction of the log odds of survival in breast cancer patients. Clinical variables such as tumor size, hormone receptor status, HER2 status, age, and treatment history were analyzed to assess their impact on survival outcomes. Data from 1557 breast cancer patients were obtained from a publicly available dataset provided by the University College Hospital, Ibadan, Nigeria. This dataset was preprocessed and analyzed using both univariate and multivariate approaches to evaluate survival outcomes. Kaplan-Meier survival curves were generated to visualize survival probabilities, while the Cox proportional hazards model identified key risk factors influencing mortality. The results showed that older age, larger tumor size, and HER2-positive status were significantly associated with an increased risk of mortality. In contrast, estrogen receptor positivity and breast-conserving surgery were linked to better survival outcomes. The findings suggest that integrating these clinical variables into predictive models improves the accuracy of survival predictions, helping to identify high-risk patients who may benefit from more aggressive interventions. This study demonstrates the potential of survival analysis in optimizing breast cancer care, particularly in resource-limited settings. Future research should focus on integrating genomic data and real-world clinical outcomes to further refine these models.

## Subject Areas

Clinical Medicine, Women's Health

## 1. Introduction

Breast cancer remains one of the most prevalent and impactful cancers worldwide, accounting for significant morbidity and mortality among women. It is widely recognized as the most common cancer affecting women, and its prevalence has increased substantially over the past few decades due to various risk factors, including genetic, hormonal, and environmental factors [1]. Approximately one in eight women will be diagnosed with invasive breast cancer during their lifetime, making it a public health priority [2].

In the context of survival analysis, log odds is a statistical measure used to express the probability of an event, such as survival, in a continuous form. This transformation is often employed in logistic regression models and facilitates a more nuanced understanding of complex relationships between variables [3]. Specifically, the log odds is the natural logarithm of the odds ratio, calculated as $\log\left(\dfrac{p}{1-p}\right)$, where $p$ represents the probability of survival. By converting survival probabilities into a continuous scale, the log odds method enhances the flexibility of survival models, allowing researchers to assess how clinical factors, such as tumor size or HER2 status, influence survival. This technique is precious in breast cancer research, where small changes in variables can significantly impact prognosis. Incorporating log odds into predictive models helps refine survival predictions, enabling more personalized treatment decisions [4].

Despite advancements in early detection, diagnosis, and treatment, predicting the prognosis of breast cancer patients remains a significant challenge. Conventional prognosis models such as the TNM classification (Tumor size, Node involvement, and Metastasis) are often limited in their ability to capture the heterogeneity of the disease, leading to suboptimal treatment decisions and patient outcomes [5]. The TNM system, though widely used, is primarily descriptive and cannot integrate various clinical and molecular data points into a cohesive risk assessment model [6]. This limitation underscores the need for more advanced statistical models capable of incorporating a wider range of clinical variables to provide more accurate prognoses for individual patients.

Breast cancer is not a homogenous disease but consists of multiple subtypes, each with distinct genetic, molecular, and clinical characteristics. These subtypes respond differently to treatment and have varying prognoses, complicating efforts to establish a universal prognostic model. For instance, hormone receptor-positive (HR+) and HER2-positive breast cancers differ significantly in their biology and treatment responses, thus requiring tailored therapeutic approaches [6]. Additionally, patient age, tumor size, and receptor status (estrogen, progesterone, and HER2) significantly affect the prognosis of breast cancer patients [7].

Given this complexity, survival analysis has emerged as a powerful tool for modeling the prognosis of breast cancer patients. Survival analysis is a statistical technique used to estimate the time until a specific event occurs, such as death, recurrence, or relapse. Unlike traditional statistical models, survival analysis accommodates censored data, which is common in medical studies where patients may be lost to follow-up or remain event-free at the time of analysis [8]. This characteristic makes it particularly suitable for breast cancer prognosis, where patient outcomes are influenced by multiple factors and often occur over extended periods.

The Cox proportional hazards model, Kaplan-Meier estimator, and parametric survival models are among the most commonly used methods in survival analysis. These models allow researchers to estimate the probability of survival over time and identify significant predictors of survival [9]. By integrating clinical variables such as tumor size, receptor status, treatment history, and patient demographics, survival analysis provides a more nuanced understanding of patient outcomes. The Cox proportional hazards model, for instance, is widely used to evaluate the effect of multiple covariates on survival time, making it an essential tool in the prognosis of breast cancer patients [10].

The use of survival analysis in breast cancer research has led to the development of predictive models that offer more personalized treatment strategies. For example, the incorporation of molecular data, such as hormone receptor and HER2 status, into survival models has improved the ability to predict patient outcomes and guide treatment decisions. In addition, parametric survival models, such as the Weibull and log-logistic models, provide flexibility in modeling survival times, particularly when the hazard function changes over time [11]. These models can handle varying survival patterns and are particularly useful in breast cancer prognosis, where the risk of recurrence or death may fluctuate depending on the stage of the disease and treatment modalities.

This study aims to build on these advancements by focusing on the log odds of survival, a transformation that converts probabilities into a continuous scale for more sophisticated statistical analysis. The log odds transformation is often used in logistic regression to model binary outcomes, such as survival versus death, and provides a more refined approach to predicting the likelihood of survival in breast cancer patients [12]. By employing survival analysis techniques, this study seeks to evaluate how key clinical variables—such as tumor size, hormone receptor status, HER2 status, and treatment history—affect the log odds of survival in breast cancer patients.

The dataset used in this study includes 1557 breast cancer patients, each with detailed clinical and demographic information. Survival analysis techniques, including the Cox proportional hazards model and parametric models, will be applied to assess the impact of clinical variables on survival outcomes. The Kaplan-Meier estimator will be used to visualize survival probabilities, while the log odds of survival will be computed to provide a more nuanced understanding of how these variables interact to influence patient prognosis [13] [14].

This research is particularly important because accurate prognostic models can significantly impact treatment decisions. Breast cancer treatment often involves a combination of surgery, chemotherapy, hormone therapy, and radiotherapy, and the choice of treatment depends heavily on the patient's prognosis. By improving the accuracy of prognostic models, clinicians can better tailor treatments to individual patients, reducing the risk of overtreatment or undertreatment [15]. For instance, patients identified as high-risk through survival analysis may benefit from more aggressive treatment, while those with favorable prognoses may avoid unnecessary interventions that could lead to adverse side effects [16].

Furthermore, survival analysis has the potential to improve healthcare resource allocation, particularly in settings where access to advanced treatments is limited. By identifying patients at higher risk, healthcare providers can ensure that resources are directed where they are most needed, ultimately improving patient outcomes and reducing healthcare costs [16]. This is particularly relevant in low-resource settings, where the efficient use of medical resources is crucial to maximizing patient care.

In conclusion, this study will use survival analysis techniques to predict the log odds of survival in breast cancer patients, focusing on key clinical variables such as tumor size, hormone receptor status, and HER2 status. By improving the accuracy of prognostic models, this research aims to contribute to the ongoing efforts to personalize breast cancer treatment and improve patient outcomes.

## 2. Review of Previous Study

The use of survival analysis in breast cancer research has gained traction due to its ability to model time-to-event data, providing valuable insights into factors influencing patient prognosis. Historically, the TNM classification system, which assesses tumor size, node involvement, and metastasis, has been the cornerstone of breast cancer prognosis. However, its limitations in accounting for disease heterogeneity have led to the development of more advanced statistical models like survival analysis [17].

One of the most frequently employed techniques in survival analysis is the Cox proportional hazards model. This model allows researchers to estimate the effect of various covariates on survival time without making specific assumptions about the baseline hazard function [18]. In breast cancer research, studies have shown that clinical factors such as age, tumor size, estrogen receptor (ER) status, and HER2 status are significant predictors of survival. For instance, larger tumor sizes and HER2 positivity are associated with increased mortality risks, while positive ER status is linked to better outcomes [19].

Additionally, the Kaplan-Meier estimator is widely used to estimate survival probabilities over time. It graphically represents survival data and provides an intuitive way to compare survival curves between different patient groups. This tool is particularly useful in understanding the impact of treatment variables, such as chemotherapy and radiotherapy, on survival outcomes [20]. Several studies have demonstrated that chemotherapy, while improving survival rates in high-risk patients, may be associated with higher mortality risks in certain populations, depending on clinical characteristics [21].

Despite its widespread use, traditional survival models like the Cox model are limited by their inability to account for unobserved heterogeneity among patients. This has led to the adoption of frailty models, which introduce random effects to capture the influence of unmeasured variables. By doing so, researchers can gain a more comprehensive understanding of patient survival trajectories and reduce bias in their estimates [22]. Frailty models have been particularly beneficial in breast cancer research, where patient responses to treatment often vary widely based on unmeasured genetic, lifestyle, and environmental factors [23].

More recent innovations in breast cancer prognosis have focused on the use of log odds transformations in survival analysis. This approach, commonly used in logistic regression, transforms survival probabilities into a continuous scale, facilitating more refined statistical analysis. Log odds models are particularly valuable in predicting binary outcomes like survival versus death, providing insights into the likelihood of survival based on specific clinical variables [23].

Parametric models, such as the Weibull and log-logistic models, have also become more popular in breast cancer research due to their flexibility in modeling hazard functions that change over time. These models are especially useful for long-term survival predictions, where the risk of recurrence or death may not remain constant throughout the follow-up period [24] [25]. For example, the log-logistic distribution is ideal for modeling survival data with initially increasing and then decreasing hazard rates—a pattern commonly observed in cancer patients [26].

In summary, survival analysis has proven to be an invaluable tool in breast cancer research, enabling more personalized treatment approaches. By integrating clinical variables such as tumor size, hormone receptor status, and treatment history into survival models, researchers can better predict patient outcomes and guide treatment decisions. However, as the complexity of breast cancer continues to evolve, so too must the statistical techniques used to model its prognosis. Future research should focus on incorporating genetic markers, lifestyle factors, and real-world clinical data into survival models to further refine their predictive power.

## 3. Method

### 3.1. Data Collection and Preparation

This study leverages a dataset from the University College Hospital, Ibadan, Nigeria, which includes detailed clinical information for 1557 breast cancer patients. The dataset was publicly accessible and provided comprehensive data on variables such as age, tumor size, hormone receptor status, HER2 status, and treatment history. Before analysis, the data were preprocessed to remove irrelevant or missing variables, ensuring accuracy in the survival analysis.

Key clinical variables were chosen for their relevance to breast cancer prognosis. These include age, tumor size, type of surgery, chemotherapy, ER status, HER2 status, and hormone therapy. The primary outcome of interest was the overall survival time, measured from diagnosis to either the event of death or censoring. This ensures that the study focuses on predicting survival probabilities with the chosen clinical factors.

### 3.2. Survival Analysis Techniques

- **Cox Proportional Hazards Model**

The Cox proportional hazards model is a semi-parametric model commonly used for survival analysis. It is particularly useful because it does not assume a specific baseline hazard function, allowing for flexibility when handling diverse datasets like breast cancer patient records. The model estimates the hazard of death, which is a function of time and various covariates (age, tumor size, ER status, etc.). The hazard function $h(t \mid X)$ for a patient is expressed as:

$$h(t \mid X) = h_0(t) \cdot \exp\left(\beta_1 X_1 + \beta_2 X_2 + \cdots + \beta_n X_n\right) \tag{1}$$

where $h_0(t)$ is the baseline hazard, $\beta$ represents the regression coefficients, and $X$ are the covariates. This model is particularly suitable for this study because it accommodates both censored data (patients who are alive or lost to follow-up) and continuous variables like age and tumor size.

- **Kaplan-Meier Estimator**

The use of the Kaplan-Meier estimator allowed for the visualization of the survival probability with time. The estimation of survival probability and the creation of survival curves are made possible by this non-parametric approach. The following represents the Kaplan-Meier estimator:

$$s(t) = \prod_{t_i \leq t} \left(1 - \frac{d_i}{n_i}\right) \tag{2}$$

where $d_i$ is the number of deaths at the time $t_i$ and $n_i$ is the number of patients at risk at a time $t_i$. Kaplan-Meier curves were generated for various subgroups (e.g., patients with HER2-positive vs. HER2-negative status) to assess how different clinical factors impact survival.

- **Parametric Models**

For robustness, parametric survival models, including the log-logistic distribution,

were applied to the data. Parametric models assume that the survival times follow a specific distribution, allowing for more precise modeling of hazard rates. The log-logistic survival function is given by:

$$S(t) = \frac{1}{1 + \left(\frac{t}{\lambda}\right)^{\gamma}}$$  (3)

where $\lambda$ is the scale parameter, and $\gamma$ is the shape parameter. This model is useful when the hazard rate initially increases and then decreases, a pattern observed in many cancer patients.

- **Model Evaluation and Selection**
- **AIC and BIC Criteria**

The Akaike Information Criterion (AIC) and the Bayesian Information Criterion (BIC) were applied to determine which model suited the data the best. Complexity models are penalized by both criteria; lower values suggest a better trade-off between fit and parsimony. Here's how these criteria are calculated:

$$AIC = 2k - 2\log(L)$$  (4)

$$BIC = log(n)k - 2\log(L)$$  (5)

where $k$ is the number of parameters, $L$ is the likelihood, and $n$ is the sample size. Models with the lowest AIC and BIC were considered optimal for survival prediction.

- **Goodness-of-Fit Tests**

The goodness-of-fit for each model was tested using the likelihood ratio test, Wald test, and score (log-rank) test. These tests assess whether the covariates significantly impact survival times. The log-rank test, for instance, evaluates whether the observed survival curves differ significantly between groups (e.g., ER-positive vs. ER-negative patients).

- **Concordance Index (C-Index)**

The predictive accuracy of each model was further evaluated using the concordance index (C-index), which measures how well the model can discriminate between patients with different survival times. A C-index value closer to 1 indicates higher predictive accuracy. Models with higher C-index values were preferred for further analysis.

- **Log Odds Transformation**

In this study, the log odds of survival were calculated as a transformation of survival probabilities, facilitating a more nuanced understanding of patient risk. The log odds function is:

$$\log\left(\frac{p}{1-p}\right) = \beta_0 + \sum_{i=1}^{n} \beta_i X_i$$  (6)

where $p$ is the probability of survival, and $X_i$ are the covariates. This transformation was used in logistic regression models to better predict binary outcomes such as survival versus death, which is particularly useful for clinical decision-making.

This methodology integrates survival analysis techniques such as the Cox proportional hazards model, Kaplan-Meier estimator, and parametric models to predict the log odds of survival in breast cancer patients. These methods, combined with rigorous model selection criteria (AIC, BIC) and evaluation techniques (C-index, log-rank test), provide a robust framework for predicting patient outcomes based on key clinical variables such as tumor size, ER status, and treatment history. The resulting models can help clinicians tailor treatment strategies, particularly in resource-limited settings where precision is critical.

## 4. Result and Finding

The dataset included 1557 breast cancer patients, and their clinical characteristics are summarized in Table 1. Table 1 highlights several significant clinical differences among breast cancer patients based on their survival outcomes. Notably, patients who died from breast cancer had a median tumor size of 25 mm, which was significantly larger than the 20 mm observed in those still alive (p < 0.001). This finding underscores the critical role that tumor size plays in survival outcomes, as larger tumors are often associated with a higher risk of mortality. Similarly, age was a key factor, with patients who died from breast cancer being younger (median age 62) compared to those who died from other causes (median age 71), indicating that younger patients with aggressive disease may face a higher mortality risk.

Additionally, the data reveal that 69% of patients who died from breast cancer underwent mastectomy, compared to 49% of survivors. This suggests that more aggressive surgical treatments, such as mastectomy, are often performed in patients with higher-risk disease, yet these interventions may not always be sufficient to improve survival in advanced cases. The higher chemotherapy rate in patients who died from breast cancer (30%) compared to survivors (23%) further supports this association with more aggressive treatment protocols for high-risk patients.

These findings emphasize the importance of early detection and personalized treatment strategies, particularly for patients with larger tumors or requiring aggressive surgical interventions.

Table 1. Clinical characteristics of breast cancer patients.

| Variable | Died of breast cancer (n = 501) | Died of other causes (n = 383) | Alive (n = 673) | P-value |
|---|---|---|---|---|
| Age (years) | 62 (51 - 71) | 71 (64 - 77) | 57 (48 - 64) | < 0.001 |
| Tumor size (mm) | 25 (20 - 35) | 23 (18 - 30) | 20 (16 - 27) | < 0.001 |
| ER Positive (%) | 72% | 87% | 76% | 0.028 |
| Mastectomy (%) | 69% | 66% | 49% | < 0.001 |
| Chemotherapy (%) | 30% | 4.4% | 23% | < 0.001 |

● Kaplan-Meier Survival Analysis

Kaplan-Meier curves were generated to estimate the survival probabilities of patients based on age, tumor size, estrogen receptor (ER) status, and HER2 status. **Figure 1** presents the survival curves for patients younger than 60 years versus those aged 60 years or older. The curve indicates that younger patients had significantly higher survival probabilities throughout the observation period (p < 0.0001). At the 5-year mark, 60% of patients under 60 were alive, compared to 42% of those over 60.

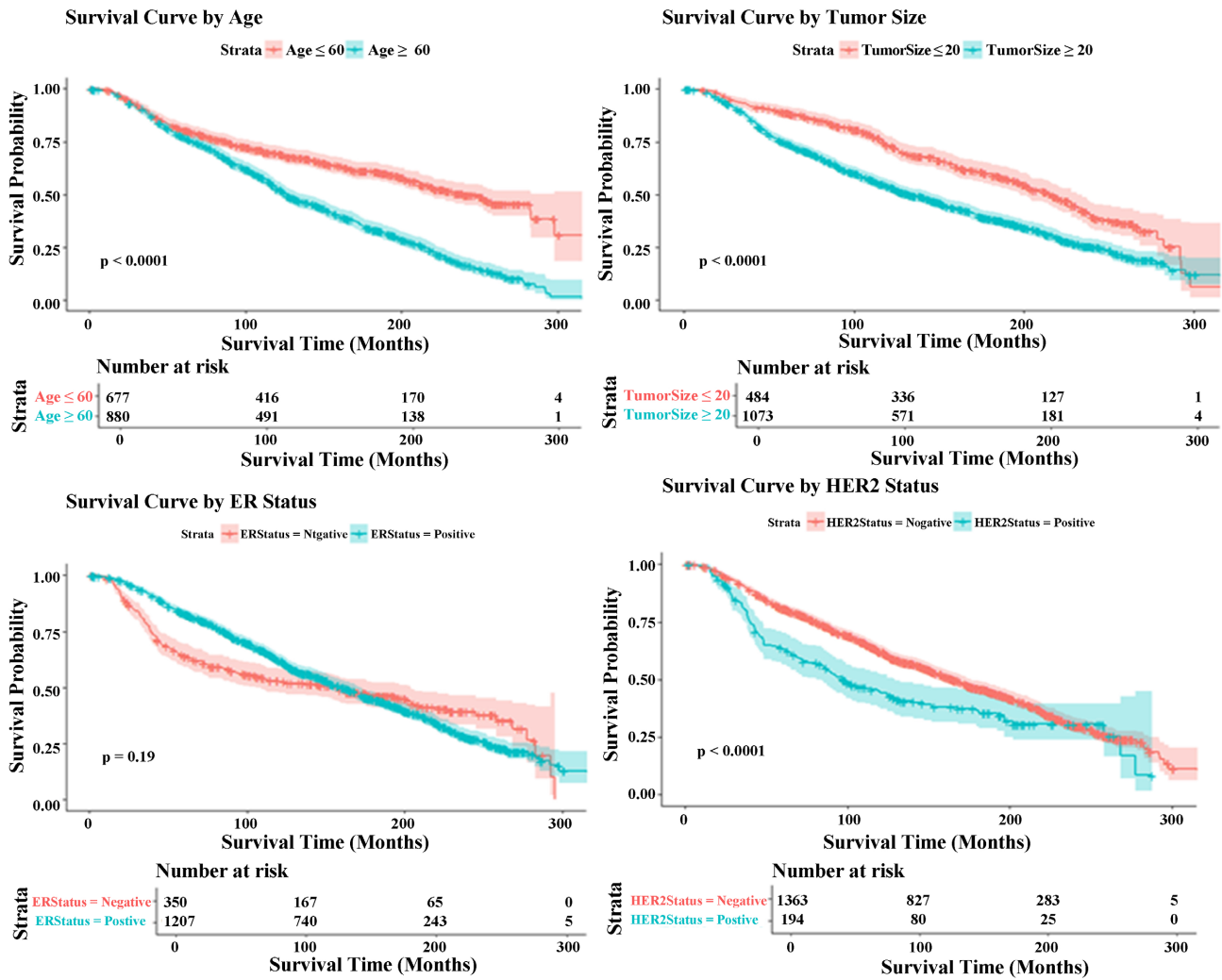

**Figure 1.** Kaplan-Meier survival curves based on clinical factors.

Similarly, survival probabilities based on tumor size showed that patients with smaller tumors (≤ 20 mm) had significantly better outcomes than those with larger tumors (> 20 mm) (p < 0.0001). By the 5-year mark, patients with smaller tumors had an 80% survival rate, while those with larger tumors had a 55% survival rate.

The Kaplan-Meier analysis also revealed minimal differences in survival between ER-positive and ER-negative patients (p = 0.19), though ER-positive patients tended to have slightly better survival outcomes. HER2 status, however, was a strong predictor of survival, with HER2-positive patients experiencing worse outcomes compared to HER2-negative patients (p < 0.001). At 5 years, 65% of HER2-positive patients had survived, compared to 78% of HER2-negative patients.

- **Cox Proportional Hazards Model**

The Cox proportional hazards model was used to evaluate the impact of various clinical variables on survival. Table 2 presents the hazard ratios for age, tumor size, ER status, HER2 status, hormone therapy, radiotherapy, chemotherapy, and type of surgery.

**Table 2.** Cox proportional hazards model.

| Variable | Coefficient | Hazard ratio (HR) | P-value | 95% CI (Lower) | 95% CI (Upper) |
|----------|-------------|-------------------|---------|----------------|----------------|
| Age (years) | 0.046 | 1.047 | < 0.001 | 1.040 | 1.054 |
| Tumor size (mm) | 0.011 | 1.011 | < 0.001 | 1.008 | 1.015 |
| ER status | −0.127 | 0.881 | 0.201 | 0.726 | 1.070 |
| HER2 status | 0.464 | 1.591 | < 0.001 | 1.296 | 1.953 |
| Hormone therapy | 0.015 | 1.016 | 0.84 | 0.874 | 1.180 |
| Radiotherapy | −0.011 | 0.989 | 0.898 | 0.834 | 1.172 |
| Chemotherapy | 0.580 | 1.786 | < 0.001 | 1.428 | 2.233 |
| Mastectomy | 0.292 | 1.339 | 0.001 | 1.122 | 1.599 |

The results show that older age and larger tumor size significantly increased the hazard of mortality. Specifically, for each additional year of age, the hazard of death increased by 4.7% (HR = 1.047, p < 0.001). Similarly, each millimeter increase in tumor size raised the hazard by 1.1% (HR = 1.011, p < 0.001). HER2-positive patients had a 59% higher risk of death compared to HER2-negative patients (HR = 1.591, p < 0.001). Chemotherapy was also associated with a significant increase in mortality risk (HR = 1.786, p < 0.001), potentially reflecting its use in higher-risk patients.

- **Log Odds of Survival**

The log odds of survival were calculated to assess the likelihood of survival versus death based on the clinical variables. Figure 2 presents the distribution of log odds, which follows a normal distribution, centered around zero. Most patients had log odds close to zero, indicating an approximately even likelihood of survival. However, patients with large tumors and HER2-positive status had negative log odds, reflecting a higher probability of death.

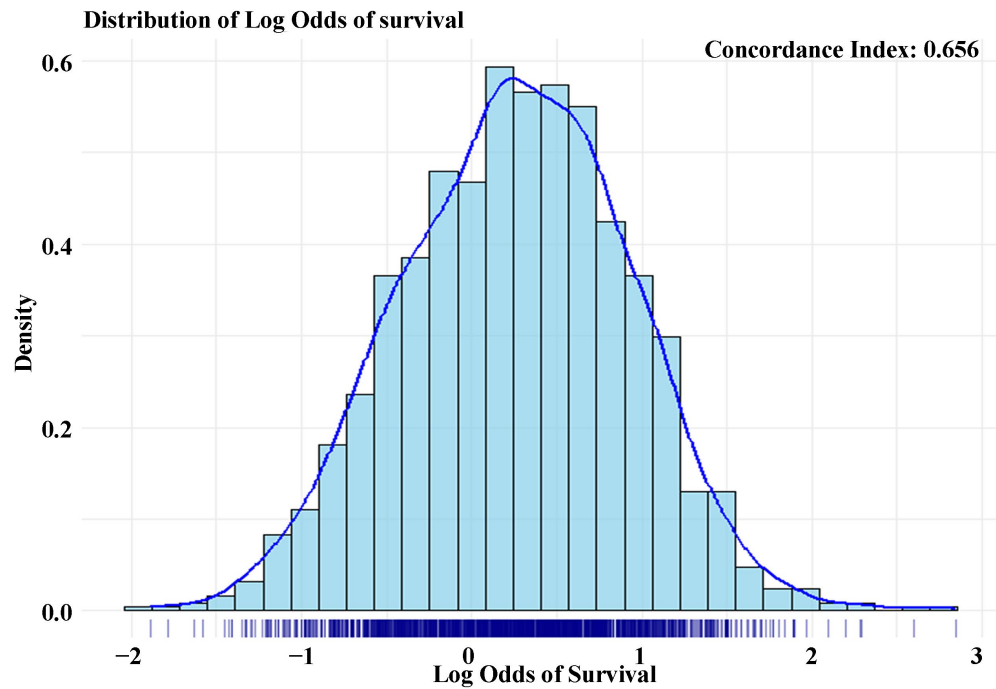

**Figure 2.** Distribution of log odds of survival.

The logistic regression model used to estimate the log odds is represented as:

$$\log\left(\frac{p}{1-p}\right) = -2.23 + 0.04 \times \textbf{\textit{Age}} + 0.0112 \times \textbf{\textit{TumorSize}} + 0.46 \times \textbf{\textit{HER2Status}} + \in$$

This model shows that age, tumor size, and HER2 status are significant predictors of survival, with older age, larger tumors, and HER2 positivity associated with worse outcomes. Each one-unit increase in tumor size results in a 12% increase in the odds of death.

- **Survival by Treatment Modality**

Survival probabilities based on different treatment modalities were also analyzed. Figure 3 presents Kaplan-Meier survival curves for patients who underwent radiotherapy, chemotherapy, and hormone therapy.

Radiotherapy was associated with improved survival outcomes, with patients receiving radiotherapy showing a significantly higher survival probability (p < 0.001). At 5 years, 70% of patients who received radiotherapy were still alive, compared to 55% of those who did not. Hormone therapy also significantly improved survival, particularly in ER-positive patients. Chemotherapy, while associated with increased mortality risk in the Cox model, showed improved outcomes in lower-risk patients who did not require more aggressive interventions.

- **Hazard Ratios and Clinical Implications**

The forest plot in Figure 4 presents hazard ratios for the key clinical variables, including age, tumor size, ER status, and HER2 status. The forest plot visually confirms the findings from the Cox model, emphasizing the significant impact of age, tumor size, and HER2 status on mortality risk.

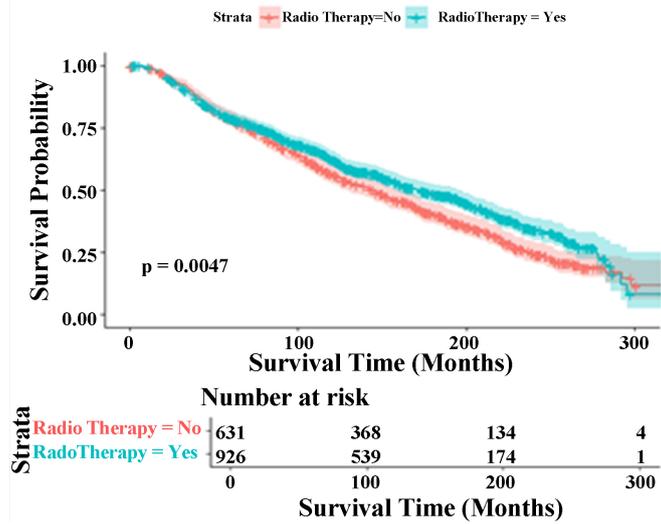

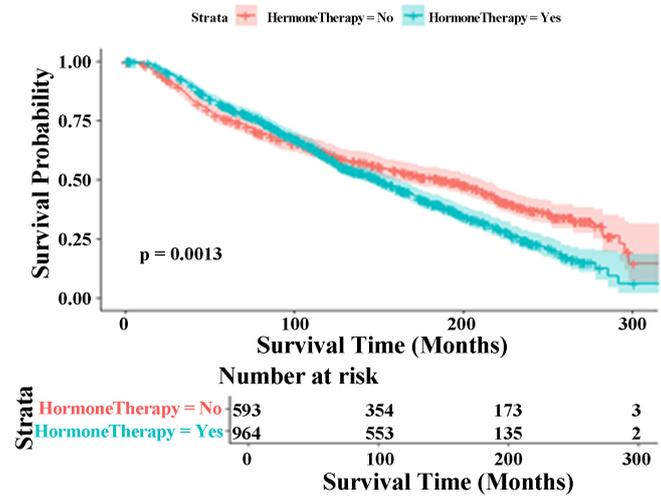

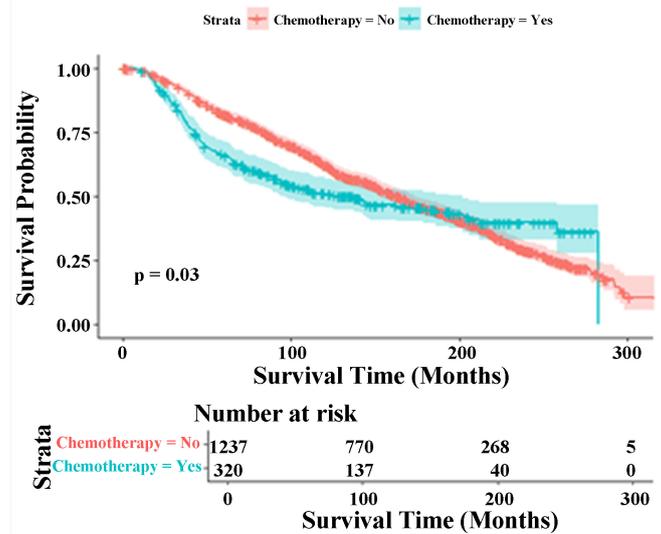

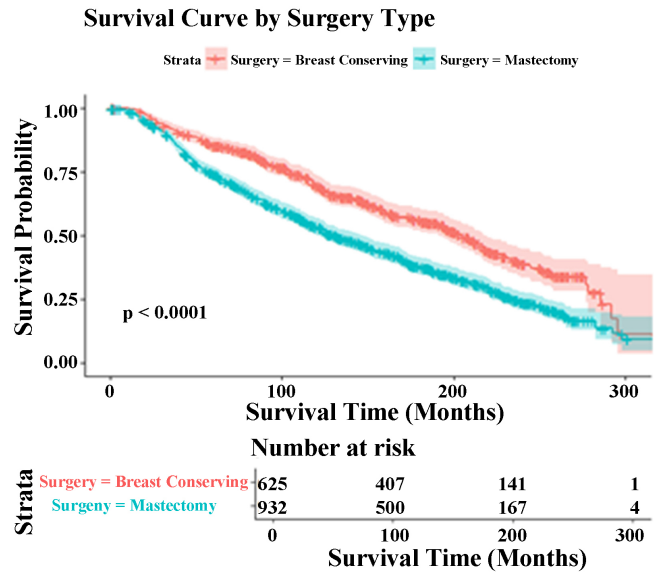

**Figure 3.** Kaplan-Meier curves based on treatment modalities.

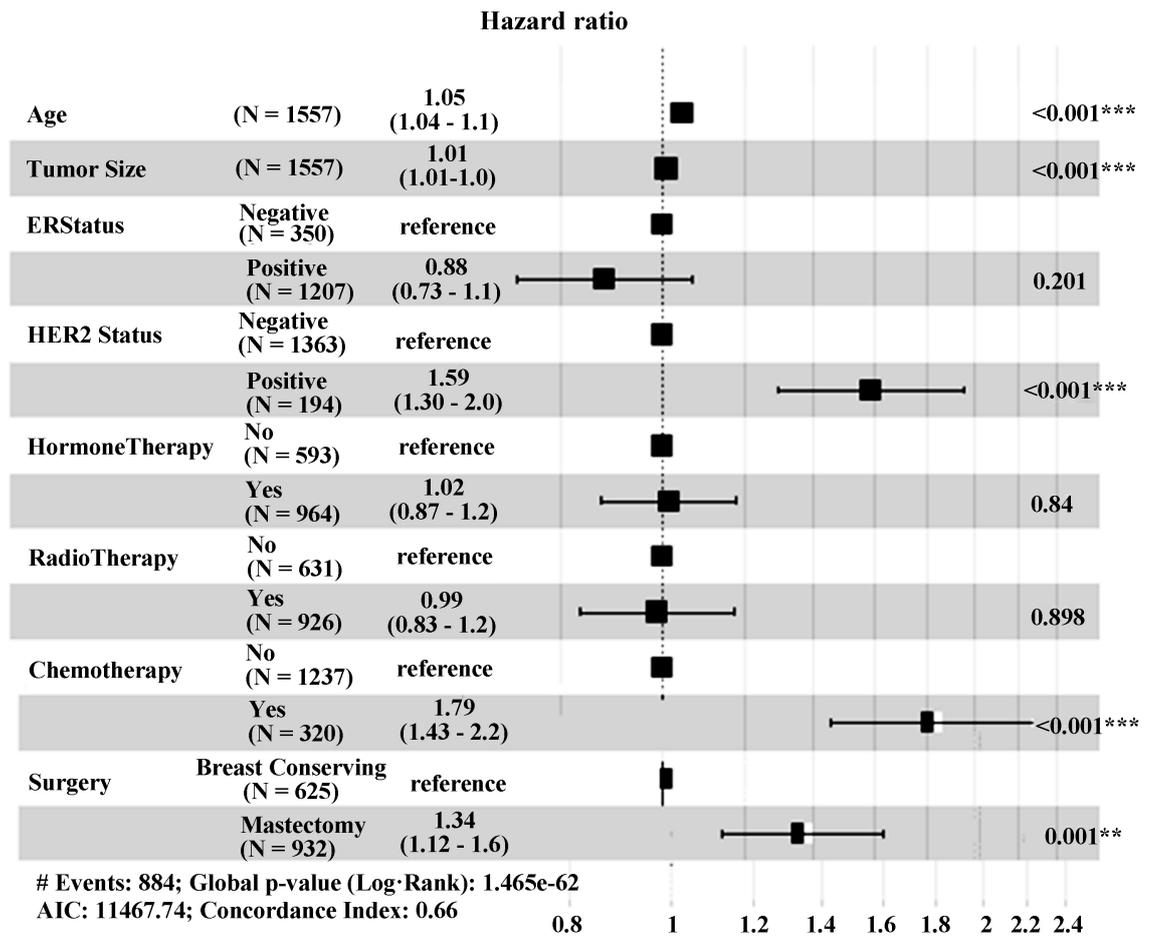

**Figure 4.** Forest plot of hazard ratios for clinical variables.

In conclusion, the analysis highlights that age, tumor size, and HER2 status are the most significant predictors of survival in breast cancer patients. Survival analysis using the Kaplan-Meier estimator and Cox proportional hazards model revealed critical insights into how these clinical factors impact patient outcomes. The findings underscore the importance of personalized treatment approaches, particularly for high-risk patients with large tumors or HER2-positive disease.

## 5. Conclusions

This study explored the survival probabilities of breast cancer patients by employing survival analysis techniques such as the Cox proportional hazards model and Kaplan-Meier estimator, focusing on clinical variables like age, tumor size, hormone receptor status (ER), HER2 status, and treatment history. The research aimed to refine the prediction of survival outcomes by assessing the impact of these factors on the log odds of survival, thereby contributing to more personalized treatment strategies for breast cancer patients.

The findings from the Cox proportional hazards model revealed that older age, larger tumor size, and HER2-positive status were significant predictors of increased mortality. Each additional year of age and millimeter of tumor size increased the hazard of death, underscoring the importance of early detection and timely treatment. HER2-positive patients faced a 59% higher risk of mortality, highlighting the need for targeted therapies for this subset of patients. In contrast, ER-positive status and hormone therapy were associated with slightly improved survival outcomes, though the effect was less pronounced than other variables.

The Kaplan-Meier survival curves further illustrated the survival differences among various patient groups. Younger patients, those with smaller tumors, and HER2-negative patients demonstrated significantly higher survival probabilities. These curves visually reinforced the statistical findings from the Cox model and provided a clear picture of how clinical variables influence survival trajectories. Additionally, treatment modalities like radiotherapy and hormone therapy were shown to improve survival, particularly in lower-risk patients, while chemotherapy, though associated with higher risk in the Cox model, still offered benefits in selected groups.

The distribution of log odds of survival further emphasized the varying risks among patients. Those with negative log odds—indicating a higher probability of death—were typically older, had larger tumors, or were HER2-positive. This transformation enabled a more nuanced understanding of survival probabilities and allowed for more precise predictions of patient outcomes based on their clinical profiles.

The implications of these findings are significant for clinical practice. By incorporating age, tumor size, and HER2 status into survival prediction models, clinicians can better identify high-risk patients who may benefit from more aggressive interventions. Conversely, patients with favorable profiles, such as smaller tumors and HER2-negative status, may avoid overtreatment and its associated side effects. This personalized approach ensures that treatment plans are tailored to individual patient risks, improving both outcomes and quality of life.

Moreover, the ability to accurately predict survival outcomes has broader implications for healthcare resource allocation, particularly in settings where access to advanced treatments is limited. By identifying patients who are most likely to benefit from specific interventions, healthcare providers can ensure that resources are used efficiently, leading to improved patient management and cost-effective care delivery.

In conclusion, this study demonstrated the value of survival analysis in predicting breast cancer outcomes. The integration of clinical variables like age, tumor size, and HER2 status into survival models significantly enhances prognostic accuracy, enabling personalized treatment strategies. These insights are crucial for improving patient outcomes and optimizing healthcare delivery, particularly for high-risk populations. Future research should explore the integration of genetic markers and real-world data to further refine these predictive models and support the ongoing efforts to personalize breast cancer care.

## Conflicts of Interes

The authors declare no conflicts of interest.